\documentclass[letterpaper]{article} 
\usepackage{aaai2027}  
\usepackage[hyphens]{url}  
\usepackage{graphicx} 
\usepackage{natbib}  
\usepackage{caption} 
\usepackage{algorithm}
\usepackage{algorithmic}

\usepackage{newfloat}
\usepackage{listings}
\DeclareCaptionStyle{ruled}{labelfont=normalfont,labelsep=colon,strut=off} 
\floatstyle{ruled}
\newfloat{listing}{tb}{lst}{}
\floatname{listing}{Listing}

\usepackage{booktabs}
\usepackage{amsmath}
\usepackage{amssymb}
\usepackage{xcolor}
\newif\ifcamerareview
\camerareviewtrue

\ifcamerareview\else\colorlet{red}{black}\fi
\usepackage{tikz}
\usepackage{multirow}
\title{%
Beyond Token-Level Cross-Entropy: Fr\'echet Distributional Post-Training for Autoregressive Image Generation
}%

\author{
    Jinhua Zhang\equalcontrib,
    Yisong Lin\equalcontrib,
    Wei Long,
    Shuhang Gu\corresponding
}
\affiliations{
    University of Electronic Science and Technology of China\\
    jinhua.zjh@gmail.com, yisongl164@gmail.com, shuhanggu@gmail.com\\
    https://github.com/CVL-UESTC/FDPT-AR
}

\begin{document}

\maketitle

\begin{abstract}

Autoregressive image generators are commonly pretrained with
token-level cross-entropy under teacher forcing, yet evaluated by the
distributional quality of decoded images. This creates an objective mismatch,
because categorical errors have unequal image-level consequences, and a
context mismatch, because inference conditions on model-generated histories.
We adapt the FD loss of Yang et al.\ to post-train a pretrained discrete generator using representation-space Fr\'echet distance as the sole objective. %
A dual-pass scheme first constructs detached rollout contexts through gradient-free generation under the model's native inference configuration, then performs differentiable replay with a probability-level straight-through estimator (STE) that reuses the sampled tokens in the forward pass while propagating image-level gradients through temperature-scaled probabilities. %
Only the generator is updated, while the tokenizer and feature extractors
remain frozen. Across eight completed configurations from four generator
families on class-conditional ImageNet at $256\times256$, FD-loss post-training
reduces FID and $\mathrm{FD}_{r6}$ by 41.4\% and 52.0\% on average. TiTok-B-64 reaches the lowest post-training FID, improving from 2.42 to 1.43 without adding parameters or inference steps. %

\end{abstract}


\section{Introduction}

\begin{figure}[t]
    \centering
    \includegraphics[width=\linewidth]{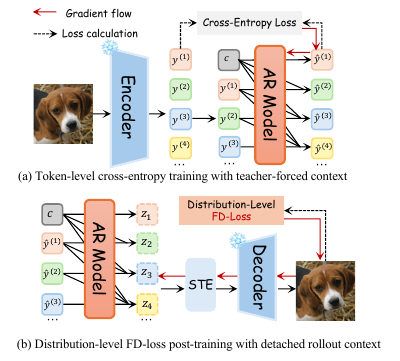}
    \caption{Teacher-forced token pretraining versus our dual-pass FD-loss
    post-training. (a) Conventional causal AR models minimize
    token-level cross-entropy using ground-truth contexts. (b) Our
    rollout pass constructs detached contexts under the model's inference configuration.
    A differentiable replay pass then applies
    a hard-forward, soft-backward probability-level STE, allowing the
    image-level FD objective to update the generator while preserving
    discrete decoding.}
    \label{fig:training-comparison}
\end{figure}

Autoregressive (AR) modeling has become a competitive paradigm for
visual generation by combining discrete image representations with scalable
Transformer architectures~\cite{dosovitskiy2021an}. In this formulation,
an image is represented as visual tokens whose conditional distributions are
predicted in a model-specific order. Advances in
tokenization~\cite{yu2021vector,lee2022autoregressive,tian2024var,weber2024maskbit,pang2026next,ren2025beyond,zhang2026mvar,shi2025scalable}
and generation order~\cite{chang2022maskgit,li2024autoregressive,pang2025randar,li2026autoregressive,huang2025spectralar}
have improved both fidelity and efficiency across token-wise, block-wise, and
scale-wise generators.

Most discrete AR generators~\cite{sun2024llamagen,yu2025randomized,tian2024var,wu2026towards}
are pretrained with token-level cross-entropy under ground-truth
contexts~\cite{huang2026self}. This paradigm creates two mismatches. First,
cross-entropy treats token prediction as categorical classification and does
not distinguish the unequal image-level consequences of different codeword
errors. Second, teacher forcing supplies ground-truth histories during
training, whereas free-running generation conditions on the model's own
outputs. Maximizing token likelihood therefore neither directly optimizes the
distribution of decoded images nor exposes the generator to its inference-time
contexts.

Image-level distribution matching provides a direct solution to address these mismatches.
In diffusion models~\cite{sun2026just,deng2026generative,geng2026improved,lu2026onestep,wang2026pixnerd,you2026advanced,li2026taming,zhou2026guiding}, Fréchet Distance (FD) loss~\cite{yang2026representation} operates within frozen feature spaces, leveraging exponential moving average (EMA)~\cite{morales-brotons2024exponential} moments to construct a stable, differentiable objective for aligning real and generated distributions.
Adapting this approach to discrete autoregressive generation, however, remains challenging.
Beyond optimization hurdles such as non-differentiable token selection, the fundamental bottleneck stems from teacher-forced conditioning.
By training the generator on ground-truth histories rather than its own predictions, teacher forcing severely exacerbates the training–inference context mismatch.
Consequently, the distribution matching loss fails to effectively optimize the model’s true inference-time rollouts, substantially undermining generation quality in practice (Table~\ref{tab:context-ablation}).

To address these challenges, we introduce an FD-loss post-training framework that couples replay on detached rollout contexts with a probability-level STE~\cite{yin2018understanding} (Figure~\ref{fig:training-comparison}).
Crucially, to bridge the teacher-forced context mismatch, our framework operates via a dual-pass scheme: the first, gradient-free rollout pass constructs detached contexts drawn entirely from the current generator under its native inference configuration. By replacing ground-truth histories with these self-generated rollouts, the generator is optimized using contexts sampled under its native inference policy.
In the second pass, the probability-level STE decodes the sampled tokens while enabling image-level FD gradients to flow back through temperature-scaled probabilities. %
Applying the EMA-based FD objective to these rollout images updates only the generator without cross-entropy, keeping the tokenizer and feature extractors strictly frozen. %

We evaluate LlamaGen~\cite{sun2024llamagen},
TiTok~\cite{yu2024image}, VAR~\cite{tian2024var}, and
GigaTok~\cite{xiong2025gigatok} on class-conditional ImageNet at
$256\times256$.
Across eight configurations, post-training reduces FID by 41.4\% and $\mathrm{FD}_{r6}$ by 52.0\% on average, without changing model size or inference steps.
TiTok-B-64 improves from an FID of 2.42 to 1.43, the lowest post-training FID among the eight configurations.
On LlamaGen-B and VAR-$d16$, replaying model-generated contexts yields lower FID than replaying ground-truth contexts (Table~\ref{tab:context-ablation}).
When we vary only the decoder readout, both models achieve their lowest FID by reusing the sampled tokens (Table~\ref{tab:readout-comparison}).
These results support distributional post-training as an image-level complement to token-level pretraining.

\begin{figure*}[!ht]
    \centering
    \includegraphics[width=\textwidth]{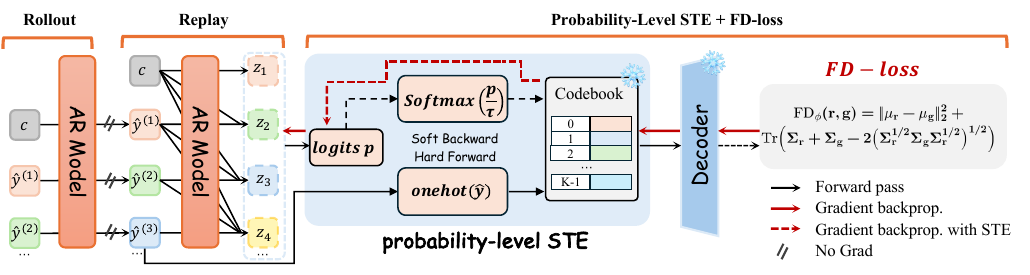}
    \caption{Overview of our FD-loss post-training framework. The dual-pass
    scheme first performs a gradient-free rollout under the checkpoint's native
    inference configuration to construct detached rollout contexts,
    and then replays these contexts to recompute codebook
    logits. A probability-level STE subsequently uses one-hot encodings of the sampled tokens for hard forward decoding while propagating the image-level FD gradient through temperature-scaled probabilities. %
    The resulting image matches the sampled rollout under identical decoding. %
    This design exposes post-training to
    model-generated contexts while bridging the gradient barrier imposed by
    discrete token selection. Only the generator is updated, without
    cross-entropy, while the tokenizer and feature extractors remain
    frozen.}
    \label{fig:framework}
\end{figure*}

\section{Related Work}

\subsection{Autoregressive Image Generation}

Discrete visual representations enable image generation through
conditional token prediction.
VQ-VAE~\cite{van2017vqvae} introduced learned
discrete representations with an autoregressive prior, and
VQGAN~\cite{esser2021taming} combined perceptually optimized tokenizers with
Transformers for high-resolution synthesis. Later work improved both token
representations~\cite{lee2022autoregressive,yu2023language,weber2024maskbit,huang2025nfig,fan2025fluid,pang2026next,ren2025beyond,han2026generative,yu2026autoregressive,lin2026ifsq} and generation order~\cite{chang2022maskgit,yu2023magvit,li2024autoregressive,pang2025randar,wang2026next}.
For example, VAR~\cite{tian2024var} predicts coarse-to-fine token scales, whereas
LlamaGen~\cite{sun2024llamagen} scales conventional next-token generation.
These systems differ architecturally but are predominantly pretrained with
token-wise cross-entropy under ground-truth contexts. We leave the tokenizer,
architecture, and generation order unchanged and instead post-train the
generator against the distribution of decoded images with
model-generated contexts.

\subsection{Learning beyond Teacher Forcing}

The discrepancy between teacher-forced and free-running
generation has motivated training under model-dependent contexts.
Scheduled Sampling~\cite{bengio2015scheduled} mixes ground-truth and predicted
inputs, Professor Forcing~\cite{lamb2016professor} aligns their hidden-state
dynamics, and Self Forcing~\cite{huang2026self} uses generated histories for
autoregressive video diffusion. For visual AR models,
RAL~\cite{ak2020incorporating} introduces free-running adversarial rewards,
whereas VA-$\pi$~\cite{liao2026va} supplies pixel-aware post-training signals.
Differentiating through discrete visual tokens is addressed by generic
straight-through estimators~\cite{yin2018understanding}, continuous
categorical relaxations~\cite{jang2017categorical}, and recent joint
tokenizer--generator training in EOSTok~\cite{chu2026endtoend} and
GEAR~\cite{lin2026gear}. RankE~\cite{jian2026ranke} further studies decoder
co-evolution under post-training.
Our setting instead freezes the tokenizer and post-trains only the
AR model. Model-generated replay reduces the teacher-forcing mismatch,
while a probability-level STE propagates the decoded-image FD objective through
discrete predictions without cross-entropy or additional inference cost.

\subsection{Distribution-Level Objectives for Generation}

Fr\'echet Inception Distance (FID) is widely used to compare real
and generated image distributions in a fixed representation space~\cite{heusel2017gans}.
Fr\'echet-GAN~\cite{doan2020image} and FastFID~\cite{mathiasen2020backpropagating} showed that related Fr\'echet objectives can also provide training gradients.
Representation Fr\'echet Loss extends this principle to EMA-estimated population statistics for post-training continuous generators~\cite{yang2026representation,sun2026just,deng2026generative,geng2026improved,lu2026onestep,wang2026pixnerd}.
Direct application to discrete generators~\cite{sun2024llamagen,yu2024image,yu2025randomized,tian2024var,xiong2025gigatok,fu2026improving}
leaves both the discrete gradient barrier and the inference-context mismatch
unresolved. Our contribution is the connection: model-generated replay places
FD optimization under free-running contexts, while a probability-level STE
preserves hard decoding and propagates the FD gradient to generator logits.

\section{Method}

We first review autoregressive generation
and representation-space Fr\'echet distance. Figure~\ref{fig:framework} then
summarizes the post-training stages: a no-gradient rollout under a model-specific inference configuration, replay under the resulting detached rollout contexts, and hard-forward, soft-backward optimization with EMA feature
moments.
While the codebook, decoder, and feature extractors remain frozen, they retain input gradients, and only the AR parameters are updated.

\subsection{Preliminaries}
\label{subsec:preliminaries}

\subsubsection{Discrete Autoregressive Image Generation}

A pretrained image tokenizer contains an encoder $\mathcal E$, a
codebook
$\mathbf C=[\mathbf e_1;\ldots;\mathbf e_K]\in\mathbb R^{K\times d}$, and a
decoder $\mathcal D$. Quantizing $\mathcal E(\mathbf x)$ represents an image
$\mathbf x$ by $T$ visual-token indices. A conditional generator with
parameters $\theta$ predicts these indices over $S$ model-specific rounds.
Let $\mathbf y^{(s)}=(y_{s,1},\ldots,y_{s,N_s})$ be the token block generated
at round $s$, where $\sum_{s=1}^{S}N_s=T$ and
$y_{s,j}\in\{1,\ldots,K\}$. For token-wise and ordered block- or scale-wise AR
models, the class-conditional distribution factorizes as
\begin{equation}
    p_\theta(\mathbf{y}\mid c)
    =\prod_{s=1}^{S}
      p_\theta(\mathbf y^{(s)}\mid\mathbf y^{(<s)},c).
    \label{eq:ar-factorization}
\end{equation}

Conventional pretraining minimizes the negative log-likelihood
\begin{equation}
    \mathcal{L}_{\mathrm{CE}}
    =-\sum_{s=1}^{S}
      \log p_\theta(\mathbf y^{(s)}\mid\mathbf y^{(<s)},c).
    \label{eq:ce}
\end{equation}

When positions within a block are conditionally independent, this
objective decomposes into token-level cross-entropy terms. Teacher forcing
provides ground-truth preceding blocks during pretraining, whereas inference
conditions on blocks produced by the model. Iterative masked generators do not
necessarily define the strict causal factorization in
Eq.~\eqref{eq:ar-factorization}. For them, $s$ indexes the native refinement
rounds, and our method operates on the corresponding discrete prediction
logits and conditioning states. Let
$\mathbf E(\hat{\mathbf y}^{(s)})
=(\mathbf e_{\hat y_{s,1}},\ldots,\mathbf e_{\hat y_{s,N_s}})$ denote the
codebook embeddings selected for a predicted block. The discrete
representation is decoded as
\begin{equation}
    \hat{\mathbf{x}}
    =\mathcal{D}\!\left(
      \mathbf E(\hat{\mathbf y}^{(1)}),\ldots,
      \mathbf E(\hat{\mathbf y}^{(S)})
    \right).
    \label{eq:hard-decode}
\end{equation}

\subsubsection{Representation-Space Fr\'echet Distance}
Let $r$ and $g$ denote real and generated image distributions,
and let $\phi$ be a frozen feature extractor. We summarize their feature
distributions by means and covariances
$(\boldsymbol{\mu}_r,\boldsymbol{\Sigma}_r)$ and
$(\boldsymbol{\mu}_g,\boldsymbol{\Sigma}_g)$. Their representation-space
Fr\'echet distance is
\begin{equation}
\begin{split}
    \operatorname{FD}_{\phi}(r,g)
    ={}&\|\boldsymbol{\mu}_r-\boldsymbol{\mu}_g\|_2^2
    +\operatorname{Tr}\big(\boldsymbol{\Sigma}_r+\boldsymbol{\Sigma}_g \\
    &-2(\boldsymbol{\Sigma}_r^{1/2}\boldsymbol{\Sigma}_g
    \boldsymbol{\Sigma}_r^{1/2})^{1/2}\big).
    \label{eq:fd}
\end{split}
\end{equation}

For the Inception-v3 representation used by the standard
evaluation protocol, Eq.~\eqref{eq:fd} corresponds to
FID~\citep{heusel2017gans}. Adapting the EMA moment estimator of
FD-loss~\citep{yang2026representation}, we precompute real moments and track
generated moments across minibatches. Historical moments are detached,
whereas current-batch moments retain gradients. We apply this estimator in
three frozen representation spaces, including the
Inception~\cite{szegedy2016rethinking}, MAE~\cite{he2022masked}, and
SigLIP~\cite{tschannen2025siglip}.

\subsection{FD-loss Post-Training}
\label{subsec:fd-loss-post-training}

We adapt only the generator $\theta$ by matching real features to those of sampled rollout images. %
Detached rollout tokens determine both the model-generated contexts and the decoded image used by the FD objective. %
A probability-level STE supplies the backward path. The codebook, decoder, and
feature extractors remain fixed but differentiable with respect to their
inputs, so the image-level gradient reaches only the generator.

\subsubsection{Base-Policy Autoregressive Rollout}

Let $\theta_0$ denote the pretrained AR parameters, and
initialize $\theta\leftarrow\theta_0$. We use
$\mathcal P_{\mathrm{base}}$ to denote the fixed model-specific inference
policy inherited from the pretrained generator, including its sampling rule,
class conditioning, classifier-free guidance (CFG)~\cite{ho2021classifierfree},
and generation order. At each post-training iteration, the current generator
produces
\begin{equation}
    \hat{\mathbf y}^{(s)}\sim
    q_{\theta,\mathcal P_{\mathrm{base}}}
    \!\left(\cdot\mid\hat{\mathbf y}^{(<s)},c\right),
    \hat{\mathbf y}
    =(\hat{\mathbf y}^{(1)},\ldots,\hat{\mathbf y}^{(S)}).
    \label{eq:rollout}
\end{equation}

Here, $q_{\theta,\mathcal P_{\mathrm{base}}}$ combines the
current parameters with the fixed decoding configuration. Gradient tracking is
disabled during rollout, and all generated indices are detached before replay.
Thus, ``base policy'' refers only to the fixed inference configuration; the
generator parameters continue to change during post-training.

\subsubsection{Detached Rollout Context Replay}

Let $\mathcal R$ denote the detached record of one rollout, including the sampled tokens and the information needed to reconstruct their prediction contexts.
For token-wise and scale-wise generators, the sampled indices determine these contexts.
For TiTok~\cite{yu2024image}, $\mathcal R$ also includes the masked states and acceptance masks from each refinement round (Table~\ref{tab:replay-map}).
The generator recomputes logits from this record using the original class condition $c$ and guidance settings.
\begin{equation}
\begin{aligned}
    \mathbf z_{1:T}
    &=\operatorname{ReplayLogits}_{\theta}\!\left(
      \mathcal I_{\mathrm{replay}}(\mathcal R),c;\right.
    \left.\mathcal P_{\mathrm{base}}\right),\\
    \mathbf z_t&\in\mathbb R^K,\qquad t=1,\ldots,T.
\end{aligned}
    \label{eq:replay}
\end{equation}

Replay therefore evaluates the generator under model-generated, rather than training-set, contexts.
The decoder reuses the hard tokens saved in $\mathcal R$, so replay requires no further sampling.
Because Eq.~\eqref{eq:rollout} is outside the computation graph,
gradients flow through the recomputed logits but not through the preceding
autoregressive decisions.

\paragraph{Hard-Forward, Soft-Backward Token Readout}

Hard token selection blocks the gradient from the decoded image
to the replay logits, whereas decoding a soft codebook mixture would alter the
forward sample. We satisfy both requirements with a probability-level STE.
For temperature $\tau>0$, we define
\begin{equation}
\begin{aligned}
    \mathbf p_t^{\mathrm{soft}}
    =\operatorname{softmax}(\mathbf z_t/\tau), 
    \quad \mathbf p_t^{\mathrm{hard}}
    =\operatorname{onehot}(\hat y_t).
\end{aligned}
    \label{eq:hard-soft-prob}
\end{equation}

Thus, $\mathbf p_t^{\mathrm{hard}}$ encodes the token sampled at the corresponding rollout position. %
We define the straight-through probability as
\begin{equation}
    \mathbf p_t^{\mathrm{st}}
    =\mathbf p_t^{\mathrm{soft}}
    +\operatorname{sg}\!\left(
      \mathbf p_t^{\mathrm{hard}}-\mathbf p_t^{\mathrm{soft}}
    \right).
    \label{eq:soft-ste}
\end{equation}

Here, $\operatorname{sg}(\cdot)$ denotes stop-gradient. By
construction, $\mathbf p_t^{\mathrm{st}}$ equals
$\mathbf p_t^{\mathrm{hard}}$ in the forward pass, while its derivative with
respect to $\mathbf z_t$ equals that of $\mathbf p_t^{\mathrm{soft}}$. The
resulting codebook embedding and decoded image are
\begin{equation}
    \widetilde{\mathbf e}_t=(\mathbf p_t^{\mathrm{st}})^\top\mathbf C,
    \qquad
    \widetilde{\mathbf{x}}
    =\mathcal{D}(\widetilde{\mathbf{e}}_1,\ldots,
    \widetilde{\mathbf{e}}_T).
    \label{eq:ste-image}
\end{equation}

The decoder thus receives only sampled codebook embeddings. 
With identical alignment and decoding, the resulting image matches the sampled rollout.
Although the codebook and decoder are frozen, their input derivatives carry
the image-level gradient to the replay logits and $\theta$.
\begin{algorithm}[t]
\caption{FD-Loss Post-Training for Discrete Image Generation}
\label{alg:ar-fd}
\begin{algorithmic}[1]
\REQUIRE Pretrained $p_{\theta_0}$; fixed base inference policy
$\mathcal P_{\mathrm{base}}$; frozen $\mathbf C$, $\mathcal D$, and
$\{\phi_m\}_{m\in\mathcal M}$; precomputed real moments; temperature $\tau$;
EMA decay $\beta$; $\epsilon>0$
\STATE Initialize $\theta\leftarrow\theta_0$
\STATE Generate an initialization set with $p_{\theta_0}$ under
$\mathcal P_{\mathrm{base}}$ and initialize detached EMA moments
\FOR{each post-training iteration}
    \STATE Sample class labels $\{c_i\}_{i=1}^{B}$
    \STATE Roll out the current $p_\theta$ under
    $\mathcal P_{\mathrm{base}}$ without gradients
    \STATE Detach each rollout and construct its architecture-specific replay
    input
    \STATE Recompute replay logits $\mathbf z_{i,1:T}$
    \STATE Construct $\mathbf p^{\mathrm{soft}}_{i,t}$, $\mathbf p^{\mathrm{hard}}$, and $\mathbf p^{\mathrm{st}}_{i,t}$ using Eqs.~\eqref{eq:hard-soft-prob}--\eqref{eq:soft-ste}
    \STATE Map $\mathbf p^{\mathrm{st}}_{i,t}$ through the frozen codebook and
    decoder, and extract $\mathbf f_i^{(m)}$ for every $m$
    \STATE Form differentiable EMA-augmented moments using
    Eq.~\eqref{eq:ema-stats}
    \STATE Compute $\mathcal L_{\mathrm{post}}$ using
    Eqs.~\eqref{eq:multi-fd}--\eqref{eq:post-loss}
    \STATE Update only $\theta$ using
    $\nabla_\theta\mathcal L_{\mathrm{post}}$; no CE term is used
    \STATE Store the detached generated EMA moments
\ENDFOR
\end{algorithmic}
\end{algorithm}
\begin{figure*}[!t]
    \centering
    \includegraphics[width=\textwidth]{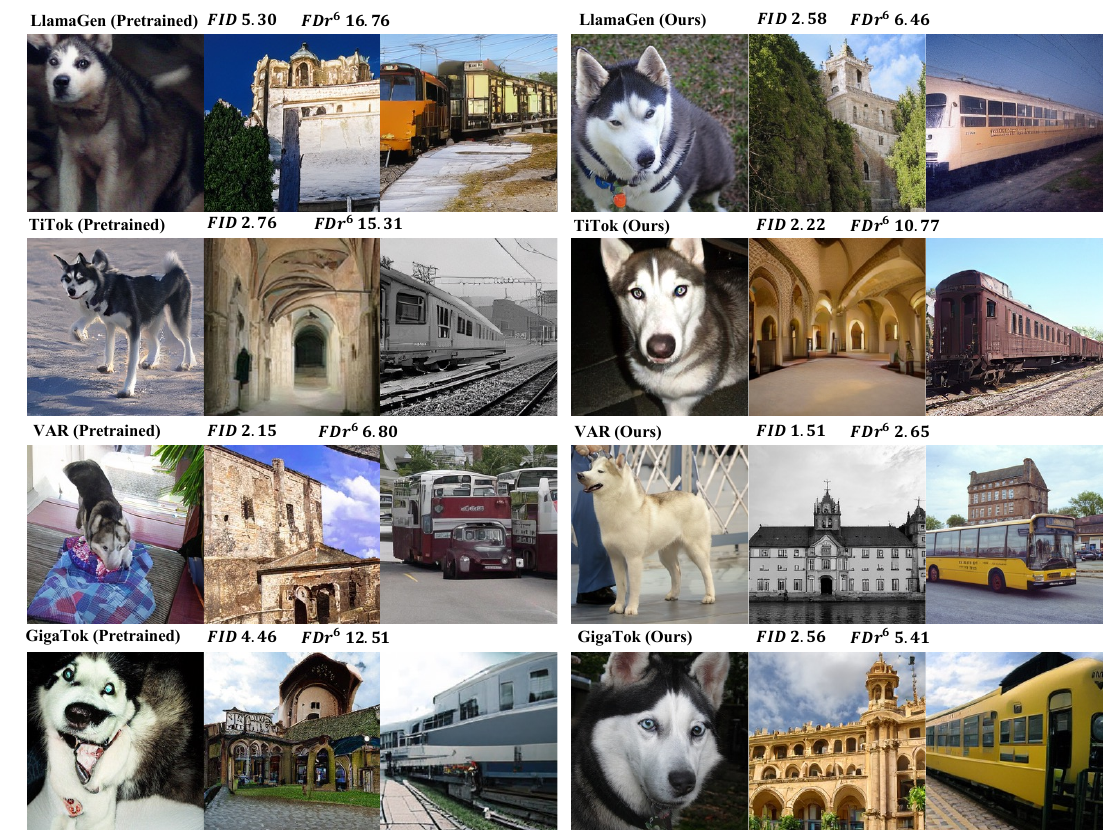}
    \caption{Qualitative comparison of pretrained generators and their
    FD-loss post-trained counterparts on class-conditional ImageNet at
    $256\times256$. Rows show LlamaGen, TiTok, VAR, and
    GigaTok-based systems. Each Base\mbox{$+$FD-loss} pair uses matched class
    conditions, random seeds, and the same model-specific inference policy.}
    \label{fig:samples_0}
\end{figure*}

\subsubsection{EMA FD Objective}

To obtain stable distribution estimates from small mini-batches, we follow \citet{yang2026representation} and maintain EMA feature statistics in multiple representation spaces. 
Let $\mathcal{M}=\{\mathrm{Inception},\mathrm{MAE},\mathrm{SigLIP}\}$ denote the
three frozen feature extractors. For each $m\in\mathcal M$, we compute
$\mathbf f_i^{(m)}=\phi_m(\widetilde{\mathbf x}_i)$ and the current-batch mean
and raw second moment
\begin{equation}
    \boldsymbol{\mu}_b^{(m)}
    =\frac{1}{B}\sum_{i=1}^{B}\mathbf{f}_i^{(m)},
    \mathbf{M}_b^{(m)}
    =\frac{1}{B}\sum_{i=1}^{B}
    \mathbf{f}_i^{(m)}(\mathbf{f}_i^{(m)})^\top.
    \label{eq:batch-moments}
\end{equation}

Before post-training, samples from the pretrained generator
$p_{\theta_0}$ under $\mathcal P_{\mathrm{base}}$ initialize detached EMA
moments $(\boldsymbol{\mu}_{\mathrm{ema}}^{(m)},
\mathbf M_{\mathrm{ema}}^{(m)})$ for each feature space. Each subsequent
iteration combines these historical statistics with differentiable
current-batch moments:
\begin{equation}
\begin{split}
    \boldsymbol{\mu}_g^{(m)}
    &=\beta\operatorname{sg}(\boldsymbol{\mu}_{\mathrm{ema}}^{(m)})
      +(1-\beta)\boldsymbol{\mu}_b^{(m)},\\
    \mathbf{M}_g^{(m)}
    &=\beta\operatorname{sg}(\mathbf{M}_{\mathrm{ema}}^{(m)})
      +(1-\beta)\mathbf{M}_b^{(m)},\\
    \boldsymbol{\Sigma}_g^{(m)}
    &=\mathbf{M}_g^{(m)}
      -\boldsymbol{\mu}_g^{(m)}(\boldsymbol{\mu}_g^{(m)})^\top.
    \label{eq:ema-stats}
\end{split}
\end{equation}

The resulting statistics aggregate information across batches
while retaining gradients through the current batch only. After updating
$\theta$, we store
$(\boldsymbol{\mu}_{\mathrm{ema}}^{(m)},
\mathbf M_{\mathrm{ema}}^{(m)})
\leftarrow
\bigl(\operatorname{sg}(\boldsymbol{\mu}_g^{(m)}),
\operatorname{sg}(\mathbf M_g^{(m)})\bigr)$
for every $m$, ensuring that no computation graph persists across
iterations. 
Rollout and replay condition on the class label $c$.
FD pools real and generated features across classes, so matching their moments does not establish that each image depicts its requested class.

Given precomputed real moments
$(\boldsymbol{\mu}_r^{(m)},\boldsymbol{\Sigma}_r^{(m)})$, we assign the same
scalar coefficient to the three raw representation-space FD terms:
\begin{equation}
    \mathcal{L}_{\mathrm{FD}}
    =\frac{1}{3}\sum_{m\in\mathcal M}
      \operatorname{FD}_{\phi_m}(r,g).
    \label{eq:multi-fd}
\end{equation}

We control the overall gradient magnitude with stop-gradient
normalization:
\begin{equation}
    \mathcal{L}_{\mathrm{post}}
    =\frac{\mathcal{L}_{\mathrm{FD}}}
    {\operatorname{sg}(\mathcal{L}_{\mathrm{FD}})+\epsilon}.
    \label{eq:post-loss}
\end{equation}

Here, $\epsilon>0$ prevents division by a vanishing loss. Because
the denominator is detached and positive, it rescales the aggregate gradient
without changing its direction. This global normalization does not normalize
the three feature-space terms separately, and equal scalar coefficients therefore
do not imply equal gradient magnitudes. Post-training minimizes only
$\mathcal{L}_{\mathrm{post}}$, and only $\theta$ is updated (Algorithm~\ref{alg:ar-fd}).
\begin{table*}[t]
\centering
\small
\setlength{\tabcolsep}{4.0pt}
\renewcommand{\arraystretch}{1.03}
\begin{tabular*}{0.98\linewidth}{
@{\extracolsep{\fill}}lcccccccc@{}
}
\toprule
Model
& Setting
& \#Params
& \#Steps
& $\mathrm{FD}_{r6}\downarrow$
& FID$\downarrow$
& IS$\uparrow$
& Prec.$\uparrow$
& Rec.$\uparrow$ \\
\midrule
\multicolumn{9}{@{}l}{\textit{Published diffusion-based generators}} \\
\addlinespace[1pt]
SiT-XL/2~\citep{ma2024sit}
& Pretrained & 675M & 200 & 8.44 & 2.12 & 256.7 & 0.81 & 0.60 \\
MAR-L~\citep{li2024autoregressive}
& Pretrained & 478M & 256 & 6.68 & 1.80 & 293.4 & 0.80 & 0.60 \\
FlowAR-H~\citep{ren2025flowar}
& Pretrained & 1.9B & 50 & 6.13 & 1.68 & 274.1 & 0.80 & 0.62 \\
MAR-H~\citep{li2024autoregressive}
& Pretrained & 942M & 256 & 5.61 & 1.56 & 299.5 & 0.80 & 0.62 \\
DeTok~\citep{yang2026latent}
& Pretrained & 478M & 256 & 5.49 & 1.39 & 306.2 & 0.81 & 0.62 \\
REG~\citep{wu2026representation}
& Pretrained & 685M & 250 & 4.64 & 1.54 & 302.9 & 0.78 & 0.62 \\
\midrule
\multicolumn{9}{@{}l}{\textit{Discrete autoregressive generators}} \\
\addlinespace[1pt]
\multirow{2}{*}{LlamaGen-B~\citep{sun2024llamagen}}
& Pretrained
& \multirow{2}{*}{111M}
& \multirow{2}{*}{256}
& 16.76 & 5.30 & 191.2 & 0.83 & 0.45 \\
& \textbf{Ours}
& & &
\textbf{6.46} & \textbf{2.58} & \textbf{273.4}
& 0.81 & 0.45 \\
\addlinespace[2pt]
\multirow{2}{*}{LlamaGen-L~\citep{sun2024llamagen}}
& Pretrained
& \multirow{2}{*}{343M}
& \multirow{2}{*}{256}
& 10.21 & 4.41 & 286.9 & 0.86 & 0.48 \\
& \textbf{Ours}
& & &
\textbf{4.07} & \textbf{1.45} & \textbf{310.2}
& 0.78 & 0.60 \\
\addlinespace[2pt]
\multirow{2}{*}{TiTok-L-32~\citep{yu2024image}}
& Pretrained
& \multirow{2}{*}{177M}
& \multirow{2}{*}{8}
& 15.31 & 2.76 & 201.7 & 0.78 & 0.57 \\
& \textbf{Ours}
& & &
\textbf{10.77} & \textbf{2.22} & \textbf{216.6}
& 0.81 & 0.57 \\
\addlinespace[2pt]
\multirow{2}{*}{TiTok-B-64~\citep{yu2024image}}
& Pretrained
& \multirow{2}{*}{177M}
& \multirow{2}{*}{8}
& 14.56 & 2.42 & 215.2 & 0.78 & 0.60 \\
& \textbf{Ours}
& & &
\textbf{9.32} & \textbf{1.43} & \textbf{249.4}
& 0.78 & 0.62 \\
\addlinespace[2pt]
\multirow{2}{*}{GigaTok-S-S~\citep{xiong2025gigatok}}
& Pretrained
& \multirow{2}{*}{111M}
& \multirow{2}{*}{256}
& 12.51 & 4.46 & 267.2 & 0.83 & 0.47 \\
& \textbf{Ours}
& & &
\textbf{5.41} & \textbf{2.56} & \textbf{289.6}
& 0.77 & 0.54 \\
\addlinespace[2pt]
\multirow{2}{*}{VAR-$d16$~\citep{tian2024var}}
& Pretrained
& \multirow{2}{*}{310M}
& \multirow{2}{*}{10}
& 11.16 & 3.32 & 274.4 & 0.84 & 0.51 \\
& \textbf{Ours}
& & &
\textbf{3.64} & \textbf{1.88} & \textbf{325.2}
& 0.79 & 0.56 \\
\addlinespace[2pt]
\multirow{2}{*}{VAR-$d20$~\citep{tian2024var}}
& Pretrained
& \multirow{2}{*}{600M}
& \multirow{2}{*}{10}
& 8.42 & 2.57 & 302.6 & 0.83 & 0.56 \\
& \textbf{Ours}
& & &
\textbf{4.77} & \textbf{1.63} & \textbf{325.5}
& 0.76 & 0.63 \\
\addlinespace[2pt]
\multirow{2}{*}{VAR-$d24$~\citep{tian2024var}}
& Pretrained
& \multirow{2}{*}{1.0B}
& \multirow{2}{*}{10}
& 6.80 & 2.15 & 330.3 & 0.82 & 0.58 \\
& \textbf{Ours}
& & &
\textbf{2.65} & \textbf{1.51} & \textbf{337.8}
& 0.80 & 0.61 \\
\bottomrule
\end{tabular*}
\caption{Quantitative comparison on class-conditional ImageNet at
$256\times256$. Published diffusion results are provided for context. Each AR
pair compares a pretrained checkpoint with our FD-loss post-training under the
same architecture and inference policy. \#Steps denotes sampling steps for
diffusion models and prediction rounds for AR models. Bold indicates the better
$\mathrm{FD}_{r6}$, FID, and IS values within each AR pair.}
\label{tab:main-results}
\end{table*}

\section{Experiments}
\subsection{Experimental Setup}
\subsubsection{Datasets and metrics}

We evaluate class-conditional generation on ImageNet-1K at
$256\times256$ using LlamaGen~\citep{sun2024llamagen},
TiTok~\citep{yu2024image}, VAR~\citep{tian2024var}, and
GigaTok~\citep{xiong2025gigatok}. These systems cover causal token-wise and
scale-wise AR generation, together with iterative discrete-token generation.
For every configuration, a public pretrained checkpoint is compared with the
same generator after FD-loss post-training.

We draw 50,000 class-balanced samples (50 per class) from each checkpoint.
We report FID~\citep{heusel2017gans}, Inception Score (IS)~\citep{Salimans2016ImprovedTF}, precision (Prec.) and recall (Rec.)~\citep{Kynknniemi2019ImprovedPA}, and $\mathrm{FD}_{r6}$~\citep{yang2026representation}.
The last metric averages normalized FD ratios across Inception-v3,
ConvNeXt-v2~\citep{woo2023convnext}, DINOv2~\citep{oquab2024dinov},
MAE~\citep{he2022masked}, SigLIP~\citep{tschannen2025siglip}, and
CLIP~\citep{radford2021learning}. Inception-v3~\citep{szegedy2016rethinking},
MAE~\citep{he2022masked}, and SigLIP define the training objective;
ConvNeXt-v2, DINOv2, and CLIP are held out from optimization. Because
$\mathrm{FD}_{r6}$ mixes optimized and held-out spaces, we interpret it as an
aggregate alignment metric rather than a fully held-out measure.

\subsubsection{Post-training details}

During post-training, we optimize only the AR generator using
Eq.~\eqref{eq:post-loss}, without retaining a cross-entropy term. The codebook, decoder, and feature extractors remain frozen, while gradients with
respect to their inputs are preserved. Unless otherwise specified, all main
experiments use a batch size of 16 for 100,000 iterations.

All eight configurations use AdamW with a learning rate of $10^{-6}$ and 1,000 warm-up steps followed by cosine decay (Table~\ref{tab:supp-common-post}).
The Inception, MAE, and SigLIP
FD terms are equally weighted. Before post-training, we initialize the EMA
feature moments using samples generated by the pretrained checkpoint and set
the decay to $\beta=0.999$. The replay temperature is fixed at $\tau=1.0$.
At each iteration, the current generator first performs a no-gradient rollout
under the checkpoint's native inference policy, including its sampling
strategy, CFG~\cite{ho2021classifierfree}, class-conditioning scheme, and
generation order.

\subsection{Main Results}

Table~\ref{tab:main-results} compares eight paired AR configurations, with
published diffusion-based generators included for context. The paired
\emph{Pretrained} and \emph{Ours} rows are evaluated under our unified
pipeline. Across all eight pairs, our post-training consistently reduces FID
by 0.54--2.96 and $\mathrm{FD}_{r6}$ by 3.65--10.30. Averaging the relative
reduction over individual pairs gives improvements of 41.4\% in FID and
52.0\% in $\mathrm{FD}_{r6}$. IS also increases for every configuration by 7.5--82.2 points, with an average gain of 32.3 points. %
The lowest resulting
FID is 1.43 on TiTok-B-64, improved from 2.42, while the lowest
$\mathrm{FD}_{r6}$ is 2.65 on VAR-$d24$.
Because the architecture, parameter count, prediction rounds, and decoding
policy remain unchanged within each pair, these comparisons isolate the
effect of FD-loss post-training under the stated protocol.

Recall rises in six of eight configurations and precision falls in six, suggesting broader coverage as FID declines.
Figure~\ref{fig:samples_0} pairs samples by class and random seed for visual comparison of sharpness and class consistency.

\subsection{Ablation Studies}
\subsubsection{Ablation protocol}

All ablations use a global batch size of 32 and 3,000 post-training iterations.
Unless otherwise stated, they are conducted on LlamaGen, and the replay-context
study in Table~\ref{tab:context-ablation} additionally evaluates VAR-$d16$ to
cover both token-wise and scale-wise AR generation. The nominal defaults are
$\beta=0.999$, $\tau=1.0$, and equal scalar coefficients for the Inception,
MAE, and SigLIP FD terms. 

\subsubsection{Replay-context source}

To examine whether the effect of replay-context source is consistent across AR
formulations, we conduct this ablation on token-wise LlamaGen and scale-wise VAR-$d16$.
For each generator, we compare the pretrained model with two
FD-loss post-training variants using either teacher-forced ground-truth
contexts or model-generated contexts. Both follow the model's native
generation order. 
Teacher-forced replay decodes predictions given real prefixes or preceding scales, while model-generated replay decodes the saved rollout. %
All other training and evaluation settings are fixed within each generator.

Table~\ref{tab:context-ablation} shows that teacher-forced replay degrades
distributional quality for both models. On LlamaGen, $\mathrm{FD}_{r6}$
increases from 16.76 to 17.07 and FID from 5.30 to 14.31. The degradation is
more pronounced on VAR-$d16$, where $\mathrm{FD}_{r6}$ increases from 11.16 to
48.77, FID from 3.32 to 43.07, and IS decreases from 274.40 to 58.51.
In contrast, model-generated replay reduces $\mathrm{FD}_{r6}$ and FID to 12.30 and 4.09 on LlamaGen, and to 6.08 and 3.14 on VAR-$d16$, respectively.
These results support model-generated replay as the more reliable context
source across the evaluated AR formulations, consistent with reducing the
context mismatch between post-training and inference.

\begin{table}[t]
\centering
\small
\setlength{\tabcolsep}{2.6pt}
\renewcommand{\arraystretch}{1.08}
\begin{tabular*}{\columnwidth}{
@{\extracolsep{\fill}}lccccc@{}
}
\toprule
Setting
& $\mathrm{FD}_{r6}\!\downarrow$
& FID$\downarrow$
& IS$\uparrow$
& Prec.$\uparrow$
& Rec.$\uparrow$ \\
\midrule
\multicolumn{6}{@{}l}{\textit{LlamaGen}~\citep{sun2024llamagen}} \\
\addlinespace[1pt]
Pretrained
& 16.76 & 5.30 & 191.24 & 0.83 & \textbf{0.45} \\
Teacher-forced
& 17.07 & 14.31 & 193.16 & 0.84 & 0.23 \\
Model-generated
& \textbf{12.30} & \textbf{4.09} & \textbf{238.78}
& \textbf{0.85} & \textbf{0.45} \\
\midrule
\multicolumn{6}{@{}l}{\textit{VAR-$d16$}~\citep{tian2024var}} \\
\addlinespace[1pt]
Pretrained
& 11.16 & 3.32 & 274.40 & \textbf{0.84} & 0.51 \\
Teacher-forced
& 48.77 & 43.07 & 58.51 & 0.39 & 0.52 \\
Model-generated
& \textbf{6.08} & \textbf{3.14} & \textbf{297.42}
& 0.80 & \textbf{0.53} \\
\bottomrule
\end{tabular*}

\caption{Replay-context ablation on LlamaGen and VAR-$d16$ after 3,000
post-training steps with batch size 32. Teacher-forced replay uses
ground-truth contexts, whereas model-generated replay uses detached rollout
contexts. Both variants are evaluated by free-running generation after post-training. %
All other settings are fixed within each generator. Bold
denotes the best result within each model block.}
\label{tab:context-ablation}
\end{table}

\subsubsection{Replay temperature}

Lowering $\tau$ sharpens the soft token probabilities while leaving the hard forward tokens fixed.
The $1/\tau$ factor in the softmax derivative also affects the scale of gradients reaching the replay logits.
As shown in
Table~\ref{tab:temperature-ablation}, every tested temperature improves
$\mathrm{FD}_{r6}$, FID, IS, and precision over the pretrained baseline, but
only $\tau=1.0$ preserves its recall. The setting $\tau=10$ achieves the lowest
$\mathrm{FD}_{r6}$ (11.12) and ties for the highest precision (0.86), but
produces a higher FID (5.08) and lower recall (0.41). In contrast,
$\tau=1.0$ achieves the lowest FID (4.09) and highest IS (238.78), preserves
the baseline recall (0.45), and reduces $\mathrm{FD}_{r6}$ to 12.30. We
therefore adopt $\tau=1.0$ as the default because it provides the most balanced
performance across distributional quality and sample coverage, although the
preferred temperature varies by metric.

\begin{table}[t]
\centering
\small
\setlength{\tabcolsep}{2.6pt}
\renewcommand{\arraystretch}{1.08}
\begin{tabular*}{\columnwidth}{
@{\extracolsep{\fill}}lccccc@{}
}
\toprule
Setting
& $\mathrm{FD}_{r6}\!\downarrow$
& FID$\downarrow$
& IS$\uparrow$
& Prec.$\uparrow$
& Rec.$\uparrow$ \\
\midrule
Pretrained
& 16.76 & 5.30 & 191.24 & 0.83 & \textbf{0.45} \\
\midrule
$\tau=0.01$
& 14.91 & 5.06 & 213.98 & \textbf{0.86} & 0.42 \\
$\tau=0.1$
& 14.33 & 4.44 & 221.49 & 0.85 & 0.43 \\
$\tau=1.0$
& 12.30 & \textbf{4.09} & \textbf{238.78}
& 0.85 & \textbf{0.45} \\
$\tau=10$
& \textbf{11.12} & 5.08 & 236.51
& \textbf{0.86} & 0.41 \\
$\tau=100$
& 11.38 & 4.99 & 235.71
& \textbf{0.86} & 0.42 \\
\bottomrule
\end{tabular*}
\caption{Replay-temperature sensitivity on LlamaGen after 3,000 post-training
steps with a global batch size of 32. Temperature affects only the soft
backward probabilities; the hard forward tokens and all other settings remain
fixed. Bold denotes the best result in each column, with ties highlighted in
all corresponding rows.}
\label{tab:temperature-ablation}
\end{table}

\begin{table}[t]
\centering
\small
\setlength{\tabcolsep}{2.2pt}
\renewcommand{\arraystretch}{1.08}
\begin{tabular*}{\columnwidth}{
@{\extracolsep{\fill}}lccccc@{}
}
\toprule
Feature spaces
& $\mathrm{FD}_{r6}\!\downarrow$
& FID$\downarrow$
& IS$\uparrow$
& Prec.$\uparrow$
& Rec.$\uparrow$ \\
\midrule
Pretrained
& 16.76 & 5.30 & 191.24 & 0.83 & 0.45 \\
\midrule
Inception
& 16.97 & \textbf{3.06} & \textbf{240.53}
& 0.78 & \textbf{0.53} \\
Inception $+$ MAE
& 14.15 & 3.20 & 239.11
& 0.81 & 0.45 \\
Inception $+$ MAE $+$ SigLIP
& \textbf{12.30} & 4.09 & 238.78
& \textbf{0.85} & 0.45 \\
\bottomrule
\end{tabular*}
\caption{Cumulative feature-space ablation on LlamaGen after 3,000
post-training steps with a global batch size of 32. The selected FD terms are
equally weighted, and all other settings are fixed. Bold denotes the best
result in each column.}
\label{tab:feature-ablation}
\end{table}

\subsubsection{Feature-space composition}

The choice of feature spaces exposes a trade-off between Inception-specific
FID and alignment across multiple representations. As shown in
Table~\ref{tab:feature-ablation}, Inception-only training achieves the lowest
FID (3.06), highest IS (240.53), and highest recall (0.53), but leaves
$\mathrm{FD}_{r6}$ nearly unchanged relative to the pretrained baseline
(16.97 versus 16.76). Adding MAE reduces $\mathrm{FD}_{r6}$ to 14.15 while
retaining an FID of 3.20 and an IS of 239.11. Using all three feature spaces
further reduces $\mathrm{FD}_{r6}$ to 12.30 and yields the highest precision
(0.85), while FID and IS remain better than the pretrained baseline. We
therefore adopt the equally weighted combination of Inception, MAE, and SigLIP
as the default, prioritizing alignment across representations over the optimum
of any single metric. Because these training representations are included in
$\mathrm{FD}_{r6}$, improvements in the aggregate score indicate alignment
with the training objective rather than fully held-out feature-space
generalization.
To evaluate alignment in the held-out spaces, we measure raw FD in ConvNeXt-v2, DINOv2, and CLIP using the same 100,000-step checkpoints as in Table~\ref{tab:main-results}.
Raw FD decreases in all three spaces for each of the eight configurations (Table~\ref{tab:heldout-cost}).
For VAR-$d16$, DINOv2 FD falls from 150.42 to 63.71 and CLIP FD from 104.01 to 46.01.

\section{Conclusion}

We introduced FD-loss post-training for pretrained discrete image
generators with frozen tokenizers and decoders. Detached rollout context replay exposes the generator to its own inference-time contexts, while a hard-forward, soft-backward probability-level STE carries decoded-image gradients across discrete token selection. %
Across eight completed ImageNet comparisons, the
method reduces FID and $\mathrm{FD}_{r6}$ by 41.4\% and 52.0\% on average
without changing inference. The evidence is bounded by the bias of the STE and
the omission of gradients through rollout trajectories, single-run ablations,
and an aggregate $\mathrm{FD}_{r6}$ metric that partly overlaps the training objective.
These results support model-generated, image-level post-training as a simple and effective complement to
token-level pretraining rather than a replacement for likelihood learning.

\section*{Acknowledgments}
This work was supported by National Natural Science Foundation
of China (No.62476051) and Sichuan Natural Science Foundation (No.2024NSFTD0041).

\bibliography{aaai2027}

\clearpage

\setcounter{table}{0}
\setcounter{figure}{0}
\renewcommand{\thetable}{S\arabic{table}}
\renewcommand{\thefigure}{S\arabic{figure}}

We first examine how EMA decay, the training objective, and the decoder readout affect post-training.
To test whether alignment improves beyond the feature spaces used for training, we compare FD before and after post-training in three held-out spaces.
Alongside these distances, we report mean update time and peak GPU memory.
We then describe how replay inputs are constructed for each architecture and document the training, inference, and evaluation settings.
We close with paired images from all eight configurations, generated before and after post-training with matched class labels, random seeds, and inference settings.

\section{Additional Ablation Studies}
\subsection{EMA-Decay Ablation}
The EMA decay controls the trade-off between responsiveness to the current
mini-batch and temporal smoothing of the generated feature moments. As shown
in Table~\ref{tab:supp-ema}, $\beta=0.99$ obtains the lowest
$\mathrm{FD}_{r6}$ (11.14) and highest precision (0.88), but increases FID
from 5.30 to 5.60 and lowers recall from 0.45 to 0.40. In contrast,
$\beta=0.999$ improves both principal distributional metrics over the
pretrained checkpoint, reducing $\mathrm{FD}_{r6}$ from 16.76 to 12.30 and
FID from 5.30 to 4.09, while increasing IS from 191.24 to 238.78.
The larger decay $\beta=0.9999$ achieves the highest IS (246.22), but yields a
less favorable FID (5.60). We therefore use $\beta=0.999$ as the default
because it provides the most balanced behavior across the reported metrics.
Because each setting is evaluated with a single short-budget run, we interpret
this experiment as a local sensitivity analysis rather than a statistical
ranking.
\begin{table}[htbp]
\centering
\small
\setlength{\tabcolsep}{3.0pt}
\renewcommand{\arraystretch}{1.06}
\begin{tabular*}{\columnwidth}{@{\extracolsep{\fill}}lccccc@{}}
\toprule
Setting
& $\mathrm{FD}_{r6}\downarrow$
& FID$\downarrow$
& IS$\uparrow$
& Prec.$\uparrow$
& Rec.$\uparrow$ \\
\midrule
Pretrained
& 16.76 & 5.30 & 191.24 & 0.83 & 0.45 \\
\midrule
$\beta=0.99$
& \textbf{11.14} & 5.60 & 235.02 & \textbf{0.88} & 0.40 \\
$\beta=0.999$
& 12.30 & \textbf{4.09} & 238.78 & 0.85 & \textbf{0.45} \\
$\beta=0.9999$
& 13.76 & 5.60 & \textbf{246.22} & 0.82 & 0.42 \\
\bottomrule
\end{tabular*}
\caption{Effect of EMA decay on LlamaGen-B after 3,000 post-training
iterations with a global batch size of 32. The pretrained checkpoint is shown
for reference. Post-training variants differ only in $\beta$, $\tau=1.0$ and
the equally weighted Inception-v3, MAE, and SigLIP FD terms remain fixed. Bold
denotes the best post-training value for each metric.}
\label{tab:supp-ema}
\end{table}

\subsection{Training Objective Ablation}
With no paired ground-truth continuation, cross-entropy on rollout tokens would train the model on its own predictions.
We therefore compare FD and maximum mean discrepancy (MMD)~\cite{JMLR:v13:gretton12a} using the same feature extractors and sampled-forward replay.
We test cubic polynomial~\cite{binkowski2018demystifying} and radial basis function (RBF)~\cite{pmlr-v37-li15} kernels.
After 3,000 updates, RBF-MMD yields the lowest $\mathrm{FD}_{r6}$ on both generators (Table~\ref{tab:objective-comparison}). %
Both MMD variants also have shorter update times and lower peak GPU memory use than FD. %
We choose FD for post-training because it achieves the lowest FID on both LlamaGen-B (4.09) and VAR-$d16$ (3.14). %

\begin{table}[htbp]
\centering
\small
\setlength{\tabcolsep}{2.25pt}
\renewcommand{\arraystretch}{1.09}
\begin{tabular*}{\columnwidth}{@{\extracolsep{\fill}}llrrrr@{}}
\toprule
Generator & Objective & $\mathrm{FD}_{r6}\downarrow$ & FID$\downarrow$ & s/update & GiB \\
\midrule
LlamaGen-B & FD              & 12.30 & \textbf{4.09} & 2.8167 & 13.87 \\
           & Polynomial-MMD  & 13.51 & 6.33 & \textbf{2.6261} & \textbf{13.51} \\
           & RBF-MMD         & \textbf{12.16} & 6.75 & 2.6437 & 13.52 \\
\midrule
VAR-$d16$  & FD              & 6.08 & \textbf{3.14} & 0.5080 & 19.40 \\
           & Polynomial-MMD  & 7.41 & 5.95 & \textbf{0.2873} & \textbf{19.03} \\
           & RBF-MMD         & \textbf{5.71} & 5.63 & 0.2905 & 19.04 \\
\bottomrule
\end{tabular*}
\caption{
Distributional objectives after 3,000 updates with a global batch size of 32. %
The last two columns report mean update time and peak allocated memory on one Pro 6000 GPU with 96 GB of memory. %
Bold marks the lowest value in each column for each generator. %
}
\label{tab:objective-comparison}
\end{table}

\subsection{Decoder Readout Ablation}
We test how the decoder input used during replay affects sample quality after post-training (Table~\ref{tab:readout-comparison}).
All three readouts use the same FD objective and replay protocol, and the two hard readouts share the same straight-through gradient rule.
At position $t$, sampled-forward STE reuses the saved rollout token $\hat y_t$, setting $\mathbf p_t^{\mathrm{hard}} = \operatorname{onehot}(\hat y_t)$.
Argmax-forward STE instead selects the token with the highest replay logit, giving $\mathbf p_t^{\mathrm{hard}} = \operatorname{onehot}(\arg\max_k z_{t,k})$.
Soft readout sets $\mathbf p_t^{\mathrm{st}} = \mathbf p_t^{\mathrm{soft}}$, so the decoder receives a probability-weighted sum of codebook embeddings at each position.
Of the three readouts, sampled-forward STE achieves the lowest FID and the highest IS on both generators under native sampling.
\begin{table}[htbp]
\centering
\small
\setlength{\tabcolsep}{3.25pt}
\renewcommand{\arraystretch}{1.09}
\begin{tabular*}{\columnwidth}{@{\extracolsep{\fill}}llccc@{}}
\toprule
Generator & Decoder readout & FID$\downarrow$ & $\mathrm{FD}_{r6}\downarrow$ & IS$\uparrow$ \\
\midrule
LlamaGen-B & Sampled-forward STE & \textbf{4.09} & \textbf{12.30} & \textbf{238.78} \\
           & Argmax-forward STE  & 12.16 & 28.81 & 150.32 \\
           & Soft readout & 6.75 & 17.50 & 224.53 \\
\midrule
VAR-$d16$  & Sampled-forward STE & \textbf{3.14} & \textbf{6.08} & \textbf{297.42} \\
           & Argmax-forward STE  & 4.46 & 15.14 & 233.69 \\
           & Soft readout & 6.94 & 19.03 & 223.99 \\
\bottomrule
\end{tabular*}
\caption{
Decoder readouts after 3,000 FD updates with a global batch size of 32.
Each model uses its native sampling policy at evaluation.
Bold marks the best value for each metric within each generator.
}
\label{tab:readout-comparison}
\end{table}

\subsection{Held-Out Feature Distances Ablation}
Because $\mathrm{FD}_{r6}$ averages normalized distances across six feature spaces, a lower score can reflect improvements in the three spaces used for training.
To test whether alignment also improves in the held-out spaces, we compare raw FD before and after post-training using ConvNeXt-v2~\cite{woo2023convnext}, DINOv2~\cite{oquab2024dinov}, and CLIP~\cite{radford2021learning} (Table~\ref{tab:heldout-cost}).
Across all eight configurations, raw FD decreases in each of the three held-out spaces.
For VAR-$d16$, for example, DINOv2 FD falls from 150.42 to 63.71 and CLIP FD from 104.01 to 46.01 after 100,000 updates.
The table also reports mean time per update and peak allocated GPU memory for each configuration.
\begin{table*}[!tp]
\centering
\small
\setlength{\tabcolsep}{3.25pt}
\renewcommand{\arraystretch}{1.09}
{
\begin{tabular*}{\textwidth}{@{\extracolsep{\fill}}lccccc@{}}
\toprule
Generator & ConvNeXt-v2 & DINOv2 & CLIP & s/update & GiB \\
\midrule
LlamaGen-B & $243.45\to163.00$ & $246.20\to174.79$ & $150.82\to125.20$ & 2.8167 & 13.87 \\
LlamaGen-L & $125.68\to83.86$ & $123.61\to59.45$ & $98.22\to55.14$ & 5.0528 & 18.64 \\
TiTok-L-32 & $189.63\to137.28$ & $189.82\to143.21$ & $158.02\to122.99$ & 0.5195 & 15.76 \\
TiTok-B-64 & $171.39\to103.99$ & $196.32\to132.66$ & $155.53\to111.39$ & 0.5080 & 14.85 \\
GigaTok-S-S & $144.32\to104.48$ & $184.29\to84.88$ & $113.78\to58.40$ & 2.2089 & 14.79 \\
VAR-$d16$ & $147.34\to117.71$ & $150.42\to63.71$ & $104.01\to46.01$ & 0.5080 & 19.40 \\
VAR-$d20$ & $110.03\to62.37$ & $100.93\to31.63$ & $85.56\to30.99$ & 0.5915 & 25.92 \\
VAR-$d24$ & $95.21\to54.86$ & $75.37\to26.59$ & $73.93\to25.02$ & 0.6932 & 34.81 \\
\bottomrule
\end{tabular*}}
\caption{
We evaluate the same eight checkpoint pairs as in Table~\ref{tab:main-results}.
Arrows show raw FD in each held-out feature space before and after 100,000 post-training updates.
The final columns give mean time per FD update and peak allocated memory on one Pro 6000 GPU with 96 GB of memory.
}
\label{tab:heldout-cost}
\end{table*}

\begin{table*}[!tp]
\centering
\small
\setlength{\tabcolsep}{3.25pt}
\renewcommand{\arraystretch}{1.09}
{
\begin{tabular}{@{}p{.16\textwidth}p{.22\textwidth}p{.31\textwidth}p{.21\textwidth}@{}}
\toprule
Generator & Saved rollout record & Input for recomputing logits & Hard decoder input \\
\midrule
LlamaGen, GigaTok & \raggedright Sampled sequence & \raggedright Class condition and right-shifted sampled sequence, with causal attention & \raggedright Saved token at each position \tabularnewline
VAR & \raggedright Sampled indices at each scale & \raggedright Class condition and quantized inputs built from all preceding scales & \raggedright Saved indices at each scale \tabularnewline
TiTok & \raggedright Masked states, acceptance masks, and final indices & \raggedright
Replay each saved masked state with its original class condition and guidance.
Use the acceptance masks to select the logits from the round in which each token was accepted.
& \raggedright Final accepted index at each position \tabularnewline
\bottomrule
\end{tabular}}
\caption{
For each architecture, the rollout record $\mathcal R$ supplies the context used to recompute logits in Eq.~\eqref{eq:replay} and the saved tokens used for hard decoding.
}
\label{tab:replay-map}
\end{table*}

\section{Implementation Details}

\subsection{Replay Inputs by Architecture}

To recompute logits during replay, we reconstruct the context in which each rollout token was selected (Table~\ref{tab:replay-map}).
For LlamaGen and GigaTok, a single causal pass over the right-shifted rollout sequence recomputes the logits at every position from its sampled prefix.
VAR applies the same principle across scales, using the sampled indices from all preceding scales to reconstruct the quantized input for the current scale.
In TiTok, a position can remain masked for several rounds, so the final token alone does not identify the context in which it was accepted.
We use the acceptance masks to identify the relevant round for each token, then recompute its logits from the masked state saved for that round.

\subsection{Configuration-Specific Training Settings}

Table~\ref{tab:supp-training-config} summarizes the architectures and baseline
pretraining settings. ``Base LR'' is the rate specified by the original
implementation. For VAR, the parenthetical value is the baseline peak after
linear scaling from a reference batch size of 256. Post-training instead uses
the optimization settings reported in Table~\ref{tab:supp-common-post}.

\begin{table*}[!tp]
\centering
\small
\setlength{\tabcolsep}{5.0pt}
\renewcommand{\arraystretch}{1.04}
\begin{tabular*}{\textwidth}{@{\extracolsep{\fill}}lcccc@{}}
\toprule
Configuration
& LlamaGen-B
& LlamaGen-L
& TiTok-L-32
& TiTok-B-64 \\
\midrule
\multicolumn{5}{@{}l}{\textit{Generator and token representation}} \\
Generator parameters
& 111M & 343M & 177M & 177M \\
Layers / width / heads
& 12 / 768 / 12 & 24 / 1024 / 16 & 24 / 768 / 16 & 24 / 768 / 16 \\
Codebook size
& 16,384 & 16,384 & 4,096 & 4,096 \\
Visual tokens
& 256 & 256 & 32 & 64 \\
Prediction rounds
& 256 & 256 & 8 & 8 \\
\midrule
\multicolumn{5}{@{}l}{\textit{Original baseline optimization}} \\
Optimizer / Adam betas
& AdamW / $(.9,.95)$ & AdamW / $(.9,.95)$
& AdamW / $(.9,.96)$ & AdamW / $(.9,.96)$ \\
Base LR
& $1{\times}10^{-4}$ & $1{\times}10^{-4}$
& $2{\times}10^{-4}$ & $2{\times}10^{-4}$ \\
LR schedule
& Constant & Constant & Cosine & Cosine \\
Warm-up / minimum LR
& – / – & – / –
& 10k / $1{\times}10^{-5}$ & 10k / $1{\times}10^{-5}$ \\
Weight decay / gradient clip
& .05 / 1.0 & .05 / 1.0 & .03 / 1.0 & .03 / 1.0 \\
Baseline training budget
& 300 epochs & 300 epochs & 500k iterations & 500k iterations \\
Baseline global batch
& 256 & 256 & 2,048 & 2,048 \\
\bottomrule
\end{tabular*}
\par\smallskip
\begin{tabular*}{\textwidth}{@{\extracolsep{\fill}}lcccc@{}}
\toprule
Configuration
& GigaTok-S-S
& VAR-$d16$
& VAR-$d20$
& VAR-$d24$ \\
\midrule
\multicolumn{5}{@{}l}{\textit{Generator and token representation}} \\
Generator parameters
& 111M & 310M & 600M & 1.0B \\
Layers / width / heads
& 12 / 768 / 12 & 16 / 1024 / 16 & 20 / 1280 / 20 & 24 / 1536 / 24 \\
Codebook size
& 16,384 & 4,096 & 4,096 & 4,096 \\
Visual tokens
& 256 & 680 & 680 & 680 \\
Prediction rounds
& 256 & 10 & 10 & 10 \\
\midrule
\multicolumn{5}{@{}l}{\textit{Original baseline optimization}} \\
Optimizer / Adam betas
& AdamW / $(.9,.95)$ & AdamW / $(.9,.95)$
& AdamW / $(.9,.95)$ & AdamW / $(.9,.95)$ \\
Base LR
& $1{\times}10^{-4}$
& $1{\times}10^{-4}$ ($3{\times}10^{-4}$)
& $1{\times}10^{-4}$ ($3{\times}10^{-4}$)
& $8{\times}10^{-5}$ ($2.4{\times}10^{-4}$) \\
LR schedule
& WSD & Linear & Linear & Linear \\
Warm-up / terminal LR ratio
& 5k / 0 & 2\% / .1 & 2\% / .1 & 2\% / .01 \\
Weight decay / gradient clip
& .05 / 1.0 & .05 / 2.0 & .05 / 2.0 & .05 / 2.0 \\
Baseline training budget
& 300 epochs & 200 epochs & 250 epochs & 350 epochs \\
Baseline global batch
& 256 & 768 & 768 & 768 \\
\bottomrule
\end{tabular*}
\caption{Architecture and original baseline optimization settings for the eight
configurations used in the main experiments. GigaTok-S-S combines the
GigaTok-B tokenizer with a 111M-parameter GPT-B generator. WSD denotes a
warm-up--stable--decay schedule. These entries describe baseline pretraining, and
the FD-loss post-training settings are given in
Table~\ref{tab:supp-common-post}.}
\label{tab:supp-training-config}
\end{table*}

\begin{table*}[!tp]
\centering
\setlength{\tabcolsep}{5.0pt}
\small
\renewcommand{\arraystretch}{1.05}
\begin{tabular*}{\textwidth}{@{\extracolsep{\fill}}lclc@{}}
\toprule
Post-training setting & Value & Post-training setting & Value \\
\midrule
Trainable module & Generator only
& Objective & $\mathcal{L}_{\mathrm{post}}$ only; no cross-entropy \\
Frozen modules
& \multicolumn{3}{l}{Tokenizer, codebook, decoder, and feature encoders} \\
Global batch & 16
& Post-training iterations & 100,000 \\
Replay temperature & $\tau=1.0$
& EMA decay & $\beta=0.999$ \\
EMA initialization & Pretrained-generator samples
& FD feature spaces & Inception-v3, MAE, SigLIP \\
FD-term coefficients & Equal
& Optimizer / LR / schedule & AdamW (.9, .95) / $1{\times}10^{-6}$ / 1k warm-up + cosine \\
\bottomrule
\end{tabular*}
\caption{Settings shared by all eight FD-loss post-training runs. All models use AdamW with betas (.9, .95), an initial learning rate of $1{\times}10^{-6}$
, 1k warm-up, and cosine decay for FD-loss post-training.}
\label{tab:supp-common-post}
\end{table*}

\subsection{Common Post-Training Protocol}
We build on the public checkpoints of
LlamaGen, TiTok, GigaTok, and VAR. Post-training updates only the generator using $\mathcal{L}_{\mathrm{post}}$. The tokenizer, codebook, and feature extractors remain frozen, and no cross-entropy term is retained. All main runs use 100,000 iterations and a global batch size of 16 using the FD post-training optimization settings reported in Table~\ref{tab:supp-common-post}. We set
$\tau=1.0$ and $\beta=0.999$, initialize the EMA moments with
pretrained-generator samples, and equally weight the Inception-v3, MAE, and
SigLIP FD terms.

\begin{table*}[!tp]
\centering
\small
\setlength{\tabcolsep}{4.0pt}
\renewcommand{\arraystretch}{1.04}
\begin{tabular*}{\textwidth}{@{\extracolsep{\fill}}lcccc@{}}
\toprule
Configuration
& LlamaGen-B
& LlamaGen-L
& TiTok-L-32
& TiTok-B-64 \\
\midrule
Generation scheme
& Causal token-wise & Causal token-wise & Iterative masked & Iterative masked \\
Generation order
& Raster & Raster & Arccos remasking & Arccos remasking \\
Prediction rounds
& 256 & 256 & 8 & 8 \\
Sampling rule
& Multinomial & Multinomial & Gumbel argmax & Gumbel argmax \\
CFG scale
& 2.0 & 2.0 & 4.5 & 3.0 \\
CFG schedule
& Constant & Constant & Linear & Linear \\
Top-$k$ / top-$p$
& All / 1.0 & All / 1.0 & N/A & N/A \\
Sampling temperature
& 1.0 & 1.0 & $9.5\!\rightarrow\!0$ & $11.0\!\rightarrow\!0$ \\
\bottomrule
\end{tabular*}
\par\smallskip
\begin{tabular*}{\textwidth}{@{\extracolsep{\fill}}lcccc@{}}
\toprule
Configuration
& GigaTok-S-S
& VAR-$d16$
& VAR-$d20$
& VAR-$d24$ \\
\midrule
Generation scheme
& Causal token-wise & Scale-wise AR & Scale-wise AR & Scale-wise AR \\
Generation order
& Raster & Next-scale & Next-scale & Next-scale \\
Prediction rounds
& 256 & 10 & 10 & 10 \\
Sampling rule
& Multinomial & Categorical & Categorical & Categorical \\
CFG scale
& 2.0 & 1.5 & 1.5 & 1.5 \\
CFG schedule
& Constant & Linear by scale & Linear by scale & Linear by scale \\
Top-$k$ / top-$p$
& All / 1.0 & 900 / .96 & 900 / .96 & 900 / .96 \\
Sampling temperature
& 1.0 & N/A & N/A & N/A \\
\bottomrule
\end{tabular*}
\caption{Native rollout and evaluation settings used by the pretrained and
post-trained generators. For TiTok, the listed temperature is the
Gumbel-randomization temperature, which is annealed across prediction rounds;
it is distinct from the replay temperature $\tau$. For VAR, the ten scales are
$1,2,3,4,5,6,8,10,13,$ and $16$.}
\label{tab:supp-inference-config}
\end{table*}

\subsection{Native Inference and Evaluation Settings}

The no-gradient rollout and final evaluation use the same model-specific
generation order and sampling policy. Consequently, post-training changes
neither the number of prediction rounds nor the inference procedure.
Table~\ref{tab:supp-inference-config} records the native quantitative
evaluation settings.
We use matched inference settings for each pair of checkpoints and generate 50,000 images from each checkpoint, with 50 samples per ImageNet class.
We use the reference feature means and covariances released by \citet{yang2026representation} for the ImageNet-1K training split (1,281,167 images).
For Inception-v3, MAE, and SigLIP, evaluation uses the same real-image statistics as the post-training objective.

In each feature space $m$, we compute the raw FD $d_m$ between the generated images and the training reference.
The corresponding FD between the 50,000 ImageNet validation images and the same reference gives the normalizer $v_m$.
Averaging the six normalized distances yields $\mathrm{FD}_{r6}=\frac{1}{6}\sum_{m=1}^{6}d_m/v_m$.
The normalizers are $1.68$ for Inception-v3, $56.87$ for ConvNeXt-v2, $14.19$ for DINOv2, $0.04$ for MAE, $0.60$ for SigLIP, and $5.60$ for CLIP.

\FloatBarrier
\section{Additional Qualitative Results}

We provide additional qualitative comparisons for all eight configurations.
For each configuration, the pretrained and post-trained generators use
identical class conditions, random seeds, and inference settings. FID and
$\mathrm{FD}_{r6}$ are reported above the corresponding samples.

\newcommand{\pairedqualfigure}[3]{%
\begin{figure*}[t]
    \centering
    \includegraphics[width=0.95\textwidth]{#1}
    \caption{Additional qualitative comparisons for #2 on class-conditional
    ImageNet at $256\times256$. The pretrained and post-trained generators
    use matched class conditions, random seeds, and inference settings.
    FID and $\mathrm{FD}_{r6}$ are shown above the corresponding samples.}
    \label{#3}
\end{figure*}
}

\pairedqualfigure
    {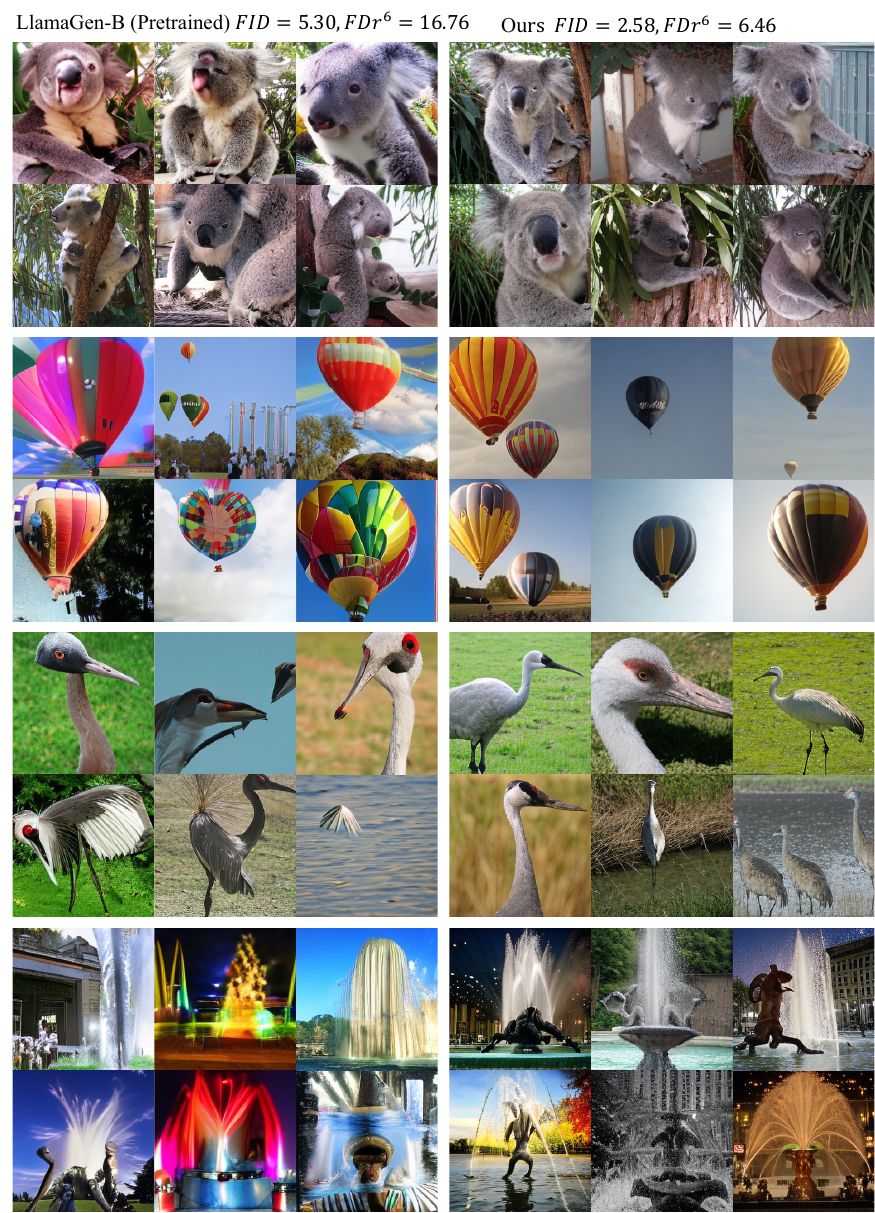}
    {LlamaGen-B}
    {fig:supp-llamagen-b}

\pairedqualfigure
    {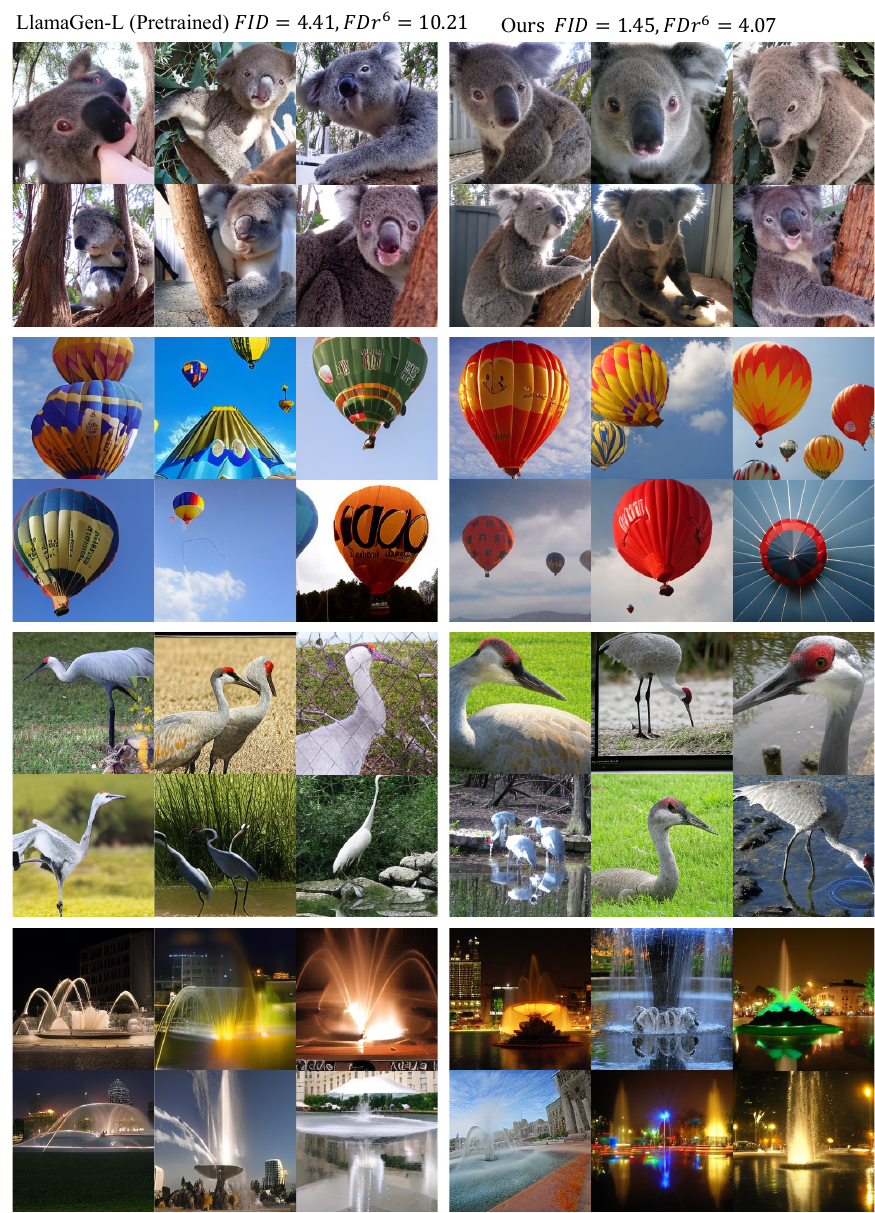}
    {LlamaGen-L}
    {fig:supp-llamagen-l}

\pairedqualfigure
    {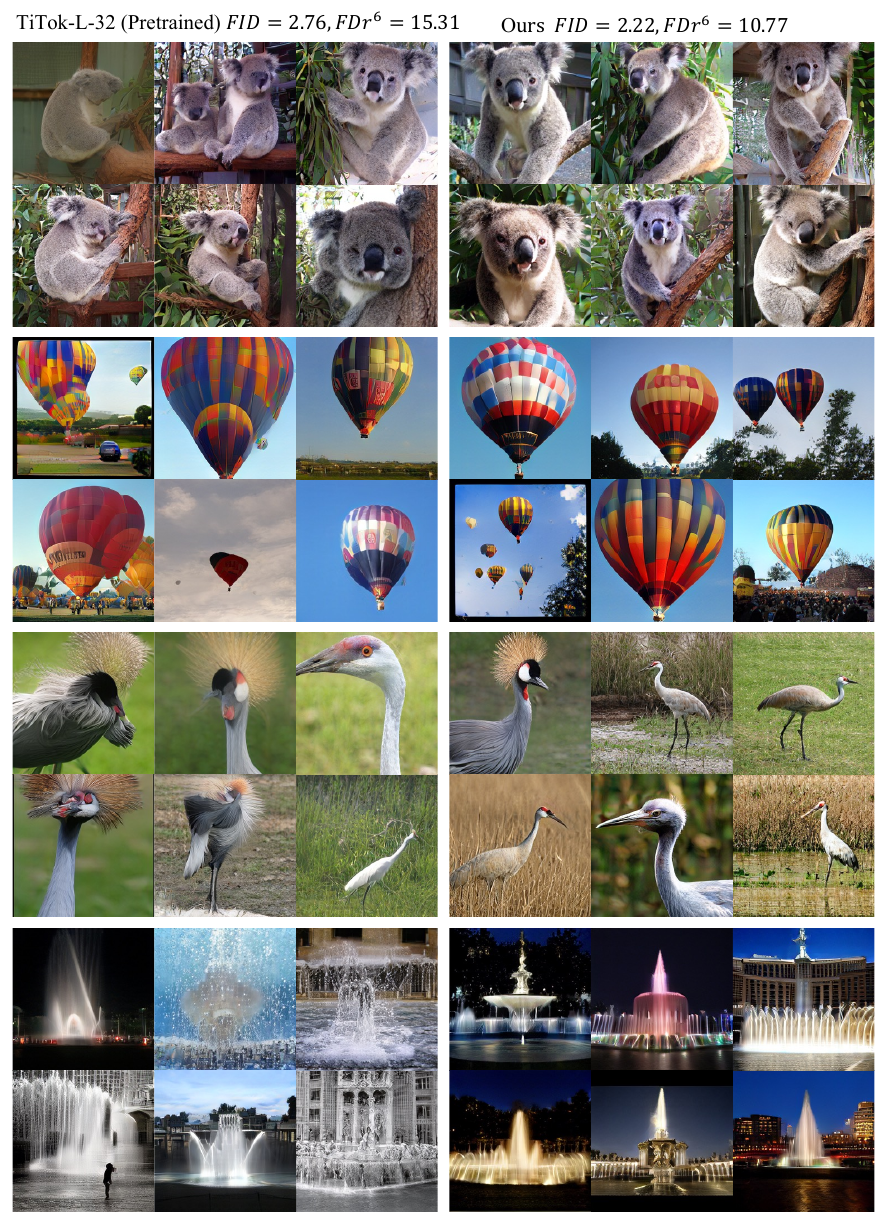}
    {TiTok-L-32}
    {fig:supp-titok-l32}

\pairedqualfigure
    {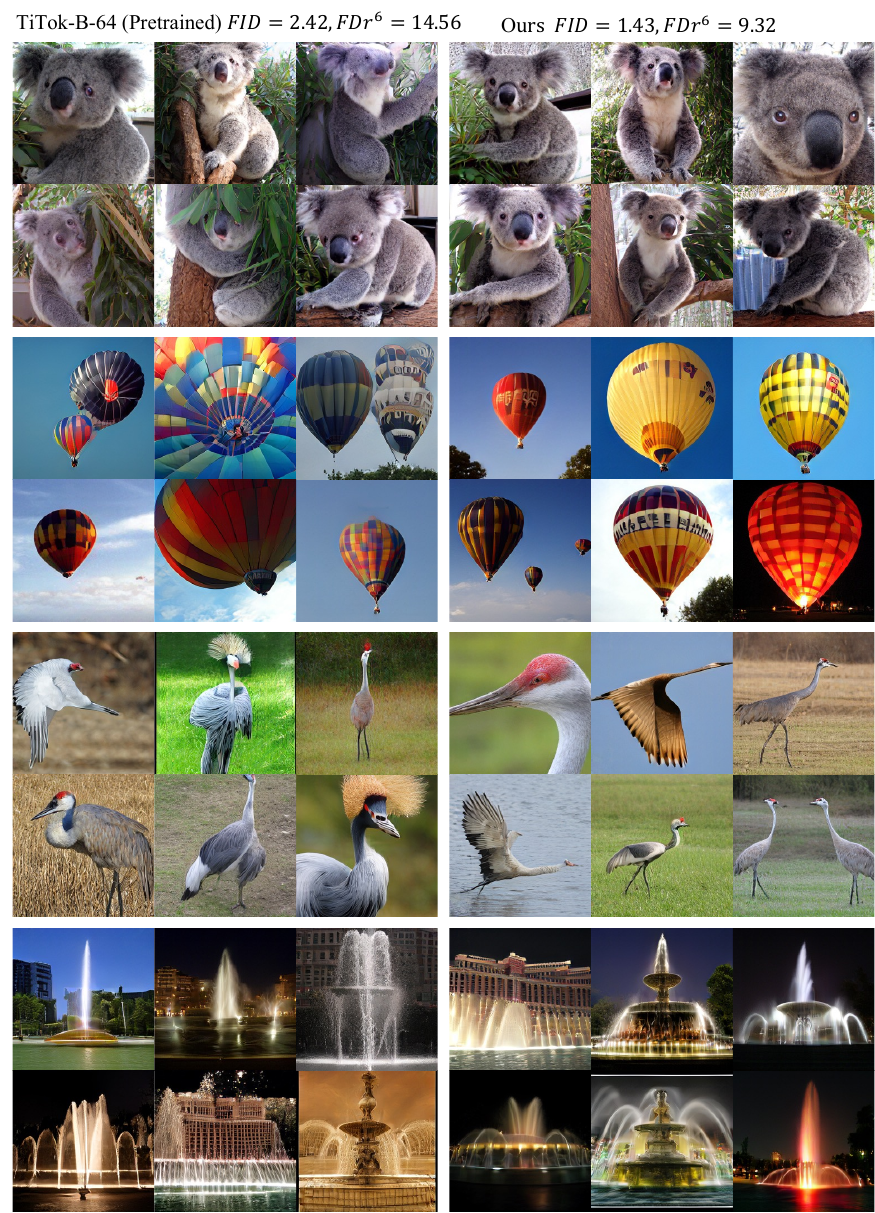}
    {TiTok-B-64}
    {fig:supp-titok-b64}

\pairedqualfigure
    {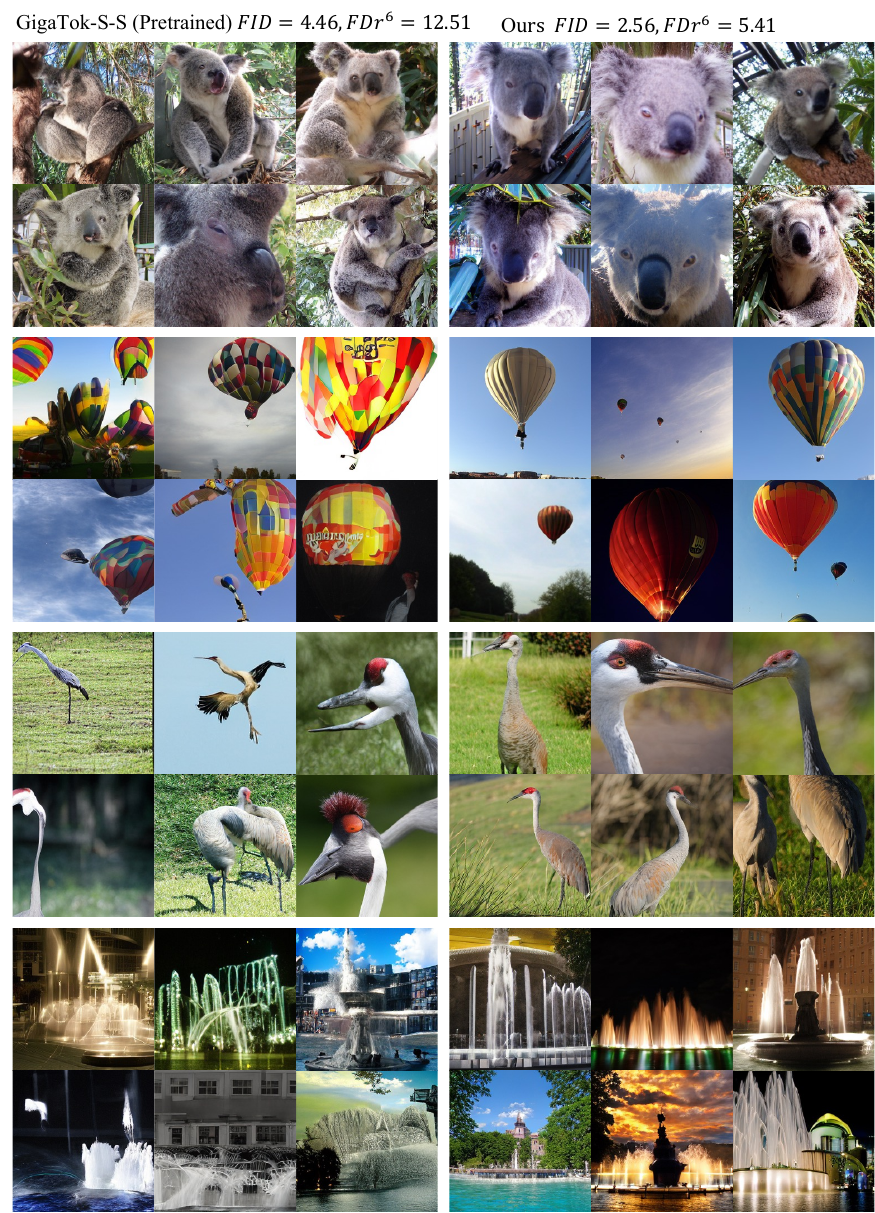}
    {GigaTok-S-S}
    {fig:supp-gigatok-bl}

\pairedqualfigure
    {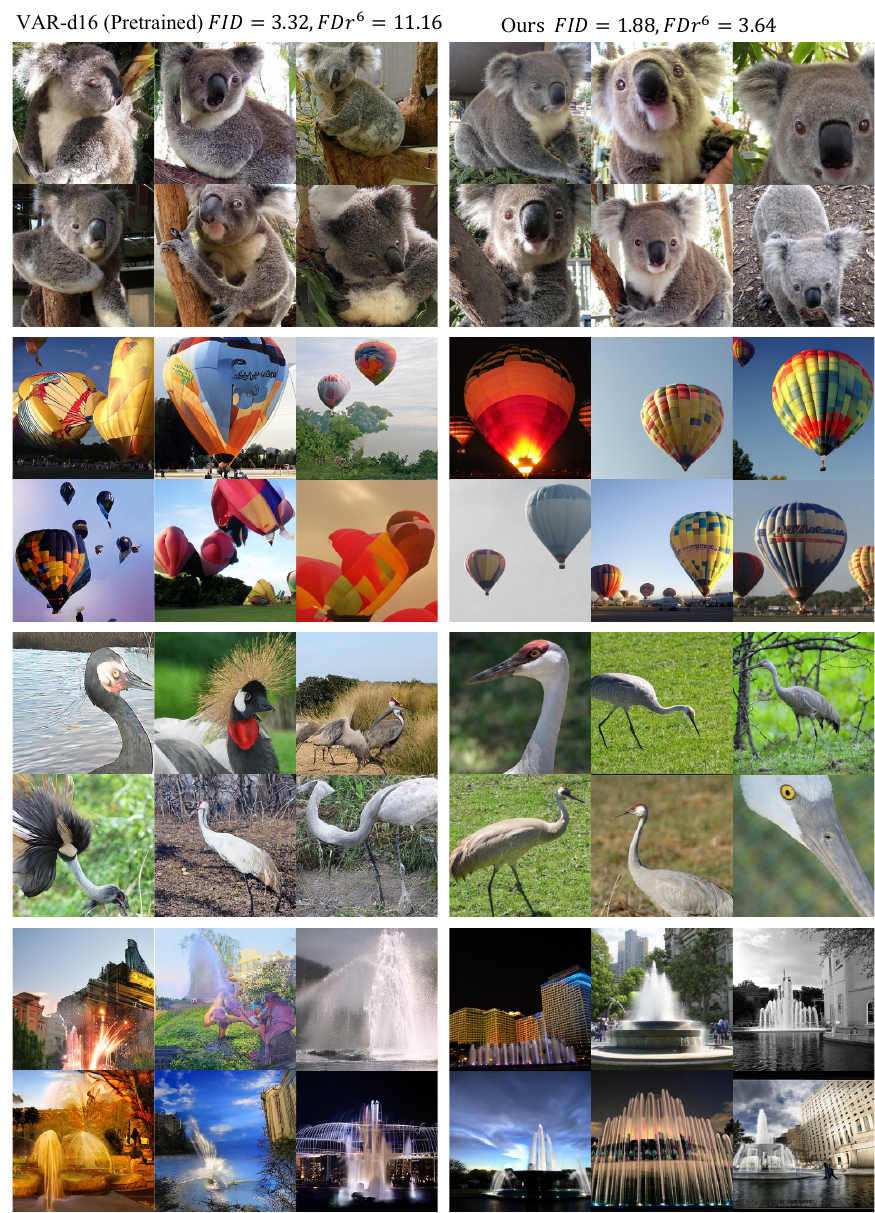}
    {VAR-$d16$}
    {fig:supp-var-d16}

\pairedqualfigure
    {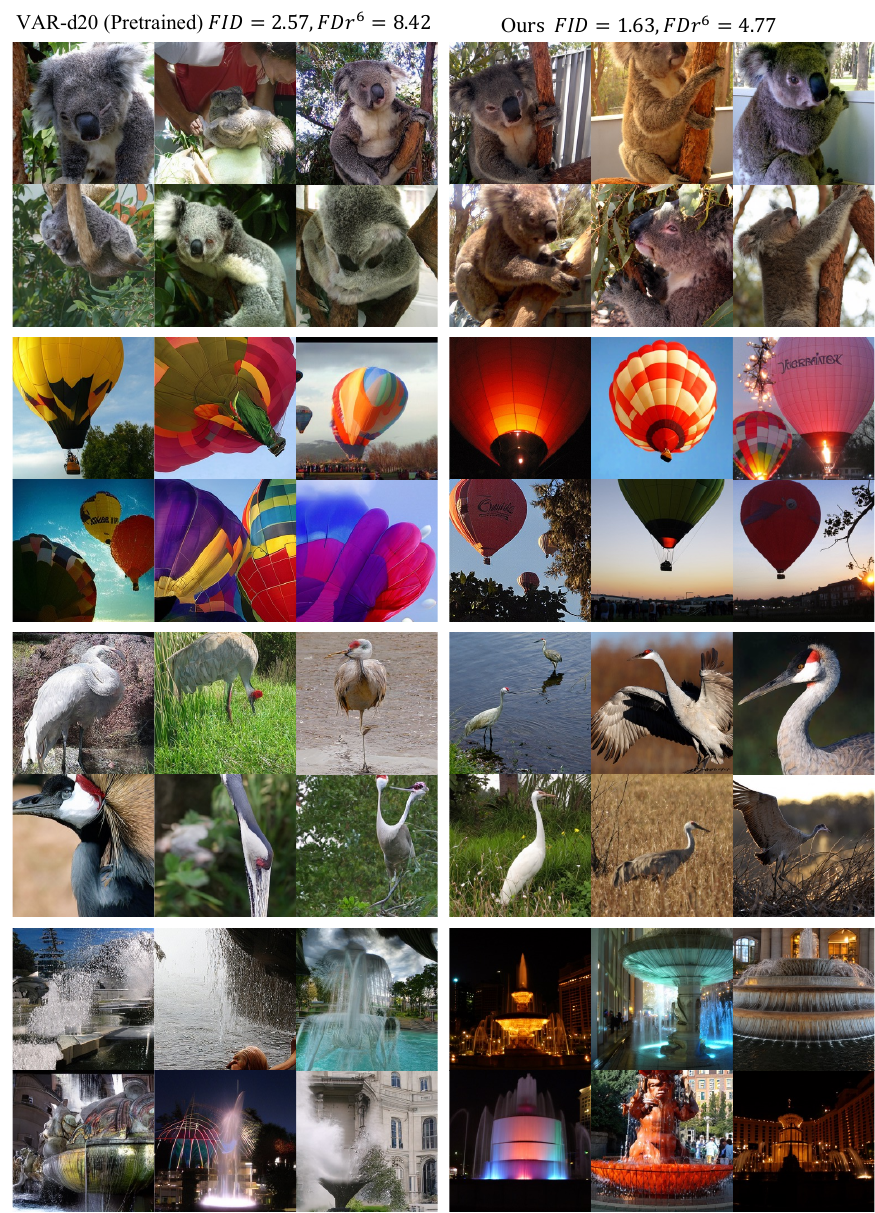}
    {VAR-$d20$}
    {fig:supp-var-d20}

\pairedqualfigure
    {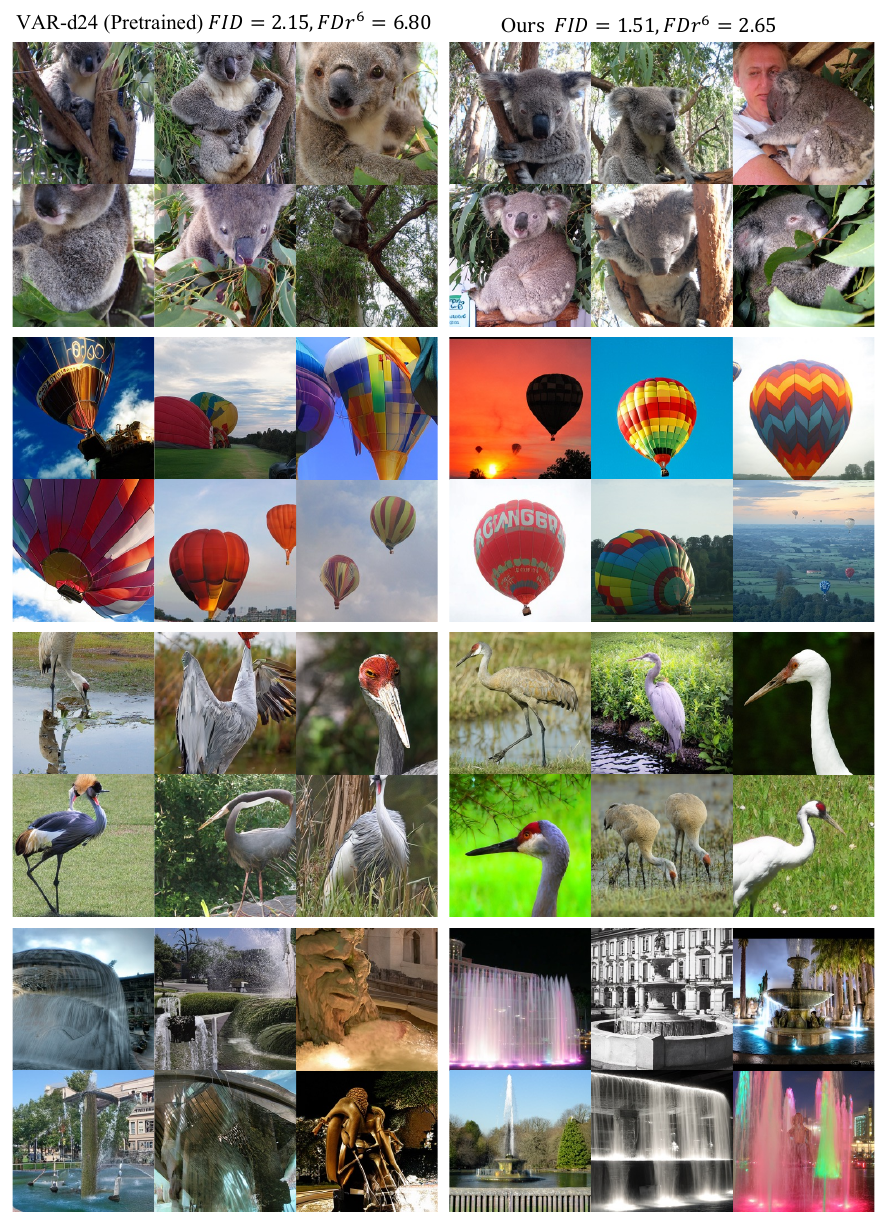}
    {VAR-$d24$}
    {fig:supp-var-d24}


\end{document}